# PERSONALIZED EXPLANATION FOR MACHINE LEARNING: A CONCEPTUALIZATION

*Research paper*

Schneider, Johannes, University of Liechtenstein, Vaduz, Liechtenstein, johannes.schneider@uni.li

Handali, Joshua Peter, University of Liechtenstein, Vaduz, Liechtenstein, joshua.handali@uni.li

## Abstract

*Explanation in machine learning and related fields such as artificial intelligence aims at making machine learning models and their decisions understandable to humans. Existing work suggests that personalizing explanations might help to improve understandability. In this work, we derive a conceptualization of personalized explanation by defining and structuring the problem based on prior work on machine learning explanation, personalization (in machine learning) and concepts and techniques from other domains such as privacy and knowledge elicitation. We perform a categorization of explainee data used in the process of personalization as well as describing means to collect this data. We also identify three key explanation properties that are amendable to personalization: complexity, decision information and presentation. We also enhance existing work on explanation by introducing additional desiderata and measures to quantify the quality of personalized explanations.*

# 1      Introduction

Techniques to extract knowledge from data, such as machine learning (ML) and related fields such as artificial intelligence, have been growing rapidly in importance over the last years. Automatic decision-making utilizing techniques from these areas increasingly support, or even replace, human decision making in many areas such as computer vision, speech recognition and natural language processing (Goodfellow et al., 2016). This is partially grounded in the emergence and improvement of complex techniques such as deep learning, which has pushed the state-of-the-art for multiple problems (Goodfellow et al., 2016). Unfortunately, complex techniques are often hard to understand for humans, earning them the title "black boxes". As a consequence, research is often driven by empirical evaluation comparing performance metrics without a thorough qualitative understanding. Furthermore, systems involving such techniques are often deemed non-trustworthy since they are susceptible to surprising errors, ie. they can be fooled in ways humans cannot (Nguyen et al., 2015). These facts underpin the need for explanations enabling a deeper understanding. Interest in explanation has also grown due to legislation, ie. the European parliament introduced the General Data Protection Regulation (GDPR) in 2018. It grants individuals the right for "meaningful explanations of the logic involved" for outcomes involving automated decision making.

To obtain meaningful and easy to understand explanations, the literature on ML has already expressed the need to focus on humans rather than just on technical aspects of ML (Adadi and Berrada, 2018; Doshi-Velez and Kim, 2017; Došilović et al., 2018; Kirsch, 2017; Ras et al., 2018). The need for explanations that are tailored to individuals (**personalized explanations**) has also been emphasized. The person (or group of persons) for whom explanations are intended (**explainee**), must provide data on herself (**explainee data**). To get meaningful **information on explainees** from raw data, sophisticated knowledge extraction and preference elicitation might be needed. Explainee data together with data used for non-personalized explanation of ML models serves as input to **personalized explanation methods**. A personalized explanation method might analyze internals of a ML model, its decision (eg. predicted class label) as well as training (and test) data in combination with explainee data and derived information thereof. It might attempt to extract a model of the explainee's decision process and align it with the ML model, eg. by identifying and comparing features found in both models.

The existing literature, however, has almost exclusively adopted the idea of personalized explanation, when personalization was part of the task performed by the ML model. For example, in recommendation systems that utilizes information on individuals such as submitted product reviews to derive predictions, the idea of personalized explanations has been expressed (Zhang et al., 2014). Thus, the idea to first collect data from the explainee, and then utilize it to improve explanations has been largely absent in the existing literature. This work attempts to close this gap by discussing both steps. We provide an overview of the current state of personalization in explanation of ML. We provide a framework covering desiderata of personalized explanations, dimensions that can be personalized, what and how information can be obtained from individuals and how this information can be utilized to customize explanations. Finally, we also discuss on how to evaluate personalized explanation methods. We do so by surveying and synthesizing existing literature on "explanation in machine learning", "personalization in machine learning" and from other domains. The structure of the paper is as follows: After laying out the methodology (Section 2), providing background information (Section 3) and conceptualizing personalized explanation (Section 4), we elaborate on explainee data in Section 5, discuss personalized explanation methods in Section 6 and, finally, present means for evaluation in Section 7.

# 2      Methodology

To develop a conceptualization, we utilized and adapted concepts, methods and ideas synthesized to a large extend in systematic literature reviews conducted by fellow researchers in ML (Explanation, Personalization and Fairness), Personalization, Privacy and Knowledge Elicitation.

Since personalized explanation in ML is an emerging field relying on many other areas, it seems suitable to a qualitative review approach such as a narrative literature review (King and He, 2005). For reproducibility a more structured approach is preferable. We adjusted the method of Webster and Watson (2002). That is, we performed forward and backward searches, derived a concept matrix, but deviated in the process of literature selection. We utilized established online databases from computer science as well as information systems such as IEEE Xplore and the AIS and ACM libraries. The development in ML is very rapid with seemingly well-renowned authors publishing their articles on the "arxiv.org" platform year or more before conference proceedings are available. Many articles are not published in journals at all. Thus, to give the reader the most up-to-date view, we included conference articles as well as articles from "arxiv.org" after careful consideration. As keywords we utilized the above listed area names (eg. personalization, explanation in ML). We limited the scope by appending "survey" and "review". For the results, we expanded the search using forward and backward search. We read titles and abstracts to filter relevant work. Our focus was not to provide a historical overview in all related areas of "Personalized explanation in ML" but to derive a conceptualization including recent developments. In particular, with respect to explanation methods in ML, which is under very rapid development, we were limited to choosing the most prominent techniques based on citations, references in surveys and suitability for personalization based on our derived concepts. Still, we point to surveys for a more detailed overview. This narrowing of articles is a limitation of our work.

## 3 Background

**Personalization** has been studied in multiple areas (Fan and Poole, 2006) such as e-commerce, computer science and cognitive sciences. Our work covers aspects from cognitive science during user modelling, ie. we make "assumptions about users' goals, interests, preferences and knowledge based on an observation of behaviour [or other sources of information]", but also from computer science, when it comes to implementing user models into an IT system, ie. we provide a platform that "supports individualized information inflow and outflow" (Fan and Poole, 2006). To personalize, information such as preferences or task knowledge must be obtained. Special **elicitation** techniques might be required, in particular to extract tacit knowledge, which is hard to express verbally or in writing (Dieste and Juristo, 2011).

**Personalization in ML** occurs in two ways: ML has been used as a means for personalization, but it has also been subject to personalization itself. Examples of the former include ML algorithms for recommender systems (Cheng et al., 2016), web personalization and search (Chen and Chau, 2004). Personalization of ML includes works on interactive ML (Amershi et al., 2014; Kulesza et al., 2015), where one seeks to improve an ML model using an iterative design process involving potentially domain experts with little ML knowledge. In contrast, personalizing explanation refers to personalizing the explanations themselves.

To "**explain**" means "to make known, to make plain or understandable" ("explain", 2018). The related term "**interpret**" can be defined as "to explain or tell the meaning of" ("interpret", 2018). **Explanation** seeks to answer questions such as: *what*, *why*, *why not*, *what if*, and *how to* (Lim et al., 2009). Gregor and Benbasat (1999) proposed the following explanation types: i) *trace or line of reasoning*, ii) *justification or support*, iii) *control or strategic*, and iv) *terminological*. The literature is not concise on using the terms. We prefer the term "explain", since we are less interested in the meaning of ML outcomes than in understanding those outcomes.

Earlier studies on **explanation in ML** such as by Huysmans et al. (2011) often investigated models deemed easily explainable such as decision trees. For more complex ML models, a large amount of techniques has been developed recently (Guidotti et al., 2018; Ras et al., 2018).

## 4 Personalized explanation in machine learning

Personalized explanations in ML refers to deriving explanations of ML models and decisions targeted to individuals as described in the introduction. To deepen the understanding, we outline desiderata of personalized explanation, introduce key concepts based on work on personalization and explanation. Then, we test the concepts for soundness and completeness by using them to characterize existing work on personalized explanation.

## 4.1 Desiderata of Personalized Explanation

Existing desiderata for explanation methods apply also to personalized explanation. Additional aspects such as privacy as well as effort for obtaining information of an explainee are also relevant:

- *Fidelity* – degree to which the explanation matches the input-output mapping of the model (Guidotti et al., 2018; Ras et al., 2018)

- *Generalizability* – range of models to which the explanation method can be applied (Ras et al., 2018)

- *Explanatory Power* – scope of questions that the explanation can answer (Ras et al., 2018)

- *Interpretability* – degree to which the explanation is human understandable (Guidotti et al., 2018). Fürnkranz et al. (2018) further distinguish an objective measure of the explanation's capability to aid the explainee in performing a task (**comprehensibility**) and a subjective measure of the explainee's acceptance of the explanation (**plausibility**).

- *Effort* – effort that an individual needs to undertake to provide additional data needed for personalization as well as the effort needed for interpretation of an explanation. The latter depends, for instance, on the complexity of the explanation. Effort for data collection only refers to data that is collected with the sole purpose of obtaining personalized explanations. Thus, if data on the explainee is already available, eg. as part of the training data for the ML model to be explained, the effort is zero. The effort for the explainee might range from answering a few simple questions to repeatedly providing feedback on proposed explanations based on careful analysis.

- *Privacy* – degree to which data on the explainee is collected, stored and used. Privacy is a key concern if information could become available to "adversaries", ie. malicious parties. In such a case, privacy might be violated even if only anonymous data is compromised (de Montjoye et al., 2015). Information on the explainee could be highly sensitive such as IQ allowing to determine a user's cognitive abilities or rather insensitive such as obtaining a user's preferred explanation method.

- *Fairness* – degree to which explanations are egalitarian (Binns, 2017; Kusner et al, 2017). While the notion of fairness is multi-faceted, a possible goal is to provide explanations of the same quality (fidelity, generalizability, interpretability, effort) for each individual

## 4.2 Conceptualization

We introduce five concepts: *explanation methods*, *personalizable explanation properties*, *data and information collection*, *personalization granularity*, and *personalization automation*. The first two concepts are based on ML explanation and the other three are on personalization. We extend Fan and Poole's (2006) categories ("What, to whom and who personalizes?") by adding a fourth dimension "How?" that describes different methods for personalized explanation.

**Personalizable explanation properties** are characteristics of an explanation that can be customizable, ie. adjusted based on explainee data. We identified three key explanation properties below.

- *Complexity* – refers to the size or number of elements of an explanation, eg. rule length or decision tree depth, and amount of relationships between features presented in an explanation, eg. correlation (Paulheim, 2012) or conjunction (Fürnkranz et al., 2018).

- *Prioritization of decision information* – refers to selection of information to present in an explanation. This include the choice of features, their relationships or examples. It is applicable on the feature space and input space. The former prioritizes by constructing an explanation using a subset of features or feature relationships. The latter refers to the subset of examples used to generate explanations. For example, a hypothetical disease diagnosis method with thousands of exemplary diagnoses of patients. To explain the diagnosis of a specific patient case to a pediatrician, a personalized explanation that uses examples might choose patient cases of children over adults.

- *Presentation* – refers to an explanation's presentation form, eg. the choice between numbers and colors to depict intensity or between natural language or logical expression to present decision rules.

The **explanation methods** can be classified according to their result (Molnar, 2018). The result of an explanation method influences how personalization may be done due to their diversity of explanation strategies and representations. The following four methods are adapted from the literature (eg. Adadi and Berrada, 2018; Lipton, 2017; Molnar, 2018; Ras et al., 2018):

- *Feature attribution* – pointing out how each feature affects the decision, eg. features importance. Feature attribution can explain the relationships between a model's intermediate components, eg. between two layers of a neural network. An attribution method points out how each contributor affects the attribution target.

- *Example-based* – returning data instances as examples to explain the model's behavior. They can be chosen from the dataset (eg. a specific target instance or representative instances) or newly created (eg. perturbed in order to serve as counterfactuals).

- *Model internals* – returning the model's internal representations, eg. structure of a decision tree, regression model, or feature visualization of a neural network.

- *Surrogate model* – returning an intrinsically interpretable, transparent model which approximates the target black-box model, eg. LIME (Ribeiro et al., 2016). This model is in turn interpreted using other explanation methods, ie. feature attribution, example-based, or model internals.

**Data and information collection** indicates how data and information from the user, in our case the explainee, is obtained. Information can refer to knowledge, eg. how a user solves the task, or to user preferences, eg. preferred colors in displays. Implicit information collection refers to information which is obtained regardless of whether explanations are needed or not. This means, the information is part of the training data of the ML model, for example in recommender systems (Zhang and Chen, 2018). In contrast, explicit information collection, information is acquired using a process that might be separate from training data collection.

**Personalization granularity** focuses on "to whom to personalize", ie. a category of individuals or a specific individual. Findings on social identity indicate (Fan and Poole, 2006) that people might behave more according to values and concerns associated with a social group in certain situations. Categorization might be a crude form of personalization, eg. we might simply categorize users into experts or non-experts, instead of assessing different dimensions related to expertise and customizing along each dimension.

**Personalization automation** focuses on "who does the personalization" (Fan and Poole, 2006), ie. manual personalization done by the explainee or automatic personalization by the system providing explanations. Manual personalization corresponds to an explainee actively setting the explanation parameters, eg. choosing the number of features to visualize.

We acknowledge that there are alternatives to our conceptualization. For instance, there are multiple options for classifying explanation methods in recent work aside from (Molnar, 2018), eg. (Adadi and Berrada, 2018; Lipton, 2017; Ras et al., 2018). One might also add a concept for "types of explanations" containing among others counterfactual and contrastive explanations (Miller, 2018). Other properties such as local and global interpretability (Guidotti et al., 2018) and explanation purpose (Ras et al., 2018) might be added. However, they might also be treated as given, ie. they might be implicit based on the task and, thus, not be subject to personalization. Furthermore, "types of explanations" can be personalized using our framework. For example, "prioritization of decision information" supports personalized counterfactual explanations by choosing counterfacts relevant to the explainee, while explanation methods such as example-based supports implementing counterfactual explanations.

## 4.3 Categorization of existing work

We assess existing work with respect to the concepts in Section 4.2. A summary is shown in Table 1.

| Personalizable Explanation Properties | | | Explanation Methods | | | | Data and Information Collection | | Personaliz. Granularity | | Personaliz. Automation | | Reference |
|---|---|---|---|---|---|---|---|---|---|---|---|---|---|
| Complexity | Prio. Features | Presentation | Feat. attribution | Example-based | Model Internals | Surr. Model | Implicit | Explicit | Individual | Category | Automatic | Manual | |
| ✓ | | | | | | ✓ | | ✓ | | (✓) | ✓ | | Lage et al., 2018 |
| ✓ | | | | ✓ | | | | | ✓ | | | ✓ | Wang et al., 2016 |
| ✓ | ✓ | | | | | ✓ | | | | (✓) | ✓ | | Ribeiro et al., 2018 |
| ✓ | ✓ | | | ✓ | | | | | | (✓) | ✓ | | Narayanan et al., 2018 |
| ✓ | ✓ | | | ✓ | | | | | | (✓) | ✓ | | Fürnkranz et al., 2018 |
| | ✓ | | ✓ | | | | ✓ | | | (✓) | ✓ | | Ross et al., 2017 |
| ✓ | | | | ✓ | | | | | | | | | Li et al., 2017 |
| ✓ | | | | | | ✓ | | | | (✓) | ✓ | | Wu et al., 2018 |
| | ✓ | ✓ | ✓ | | | | ✓ | | ✓ | | ✓ | | Chen et al., 2018 |
| | ✓ | ✓ | ✓ | | | | ✓ | | ✓ | | ✓ | | Chang et al., 2016 |
| | ✓ | | ✓ | | | | ✓ | | ✓ | | ✓ | | Zhang et al., 2014 |
| | ✓ | ✓ | ✓ | | ✓ | | | ✓ | ✓ | | | ✓ | Olah et al., 2018 |
| ✓ | | ✓ | | | | ✓ | | | | (✓) | | ✓ | Ribeiro et al., 2016 |
| ✓ | | | | ✓ | | | | | | ✓ | ✓ | | Lim et al., 2009 |
| | ✓ | | ✓ | | | | | | | (✓) | ✓ | | Dhurandhar et al., 2018 |
| ✓ | ✓ | ✓ | ✓ | | | | | ✓ | ✓ | | | ✓ | Sokol and Flach, 2018 |
| | ✓ | ✓ | ✓ | | | | ✓ | | | | | | Zhang et al., 2018 |
| ✓ | | | ✓ | ✓ | | | | | ✓ | | ✓ | | Quijano-Sanchez et al., 2017 |
| ✓ | | | ✓ | | | | | | | | | | Lundberg and Lee, 2017 |
| ✓ | | | | ✓ | | | | | | | | ✓ | Koh and Liang, 2017 |
| ✓ | | ✓ | ✓ | | | | | | | | | ✓ | Montavon et al., 2018 |

*Table 1.   Prior work categorized using concepts for personalization in ML. The ✓ symbol indicates that a concept applies. (✓) indicates that a paper only addresses one member of a categorization.*

With respect to personalizable explanation properties, two operationalisations are observed for complexity, namely size and interaction between features (Fürnkranz et al., 2018; Narayanan et al., 2018). The former refers to measures such as rule length or tree depth, while the latter refers to interactions such as disjunctions or conjunctions. Adjustments on prioritized features are mainly done on explanations for recommender systems, except for the work presented in (Ross et al., 2017). As for adjustments on presentation, prior work compared explanation presentations such as textual with graphical (Chen et al., 2018; Quijano-Sanchez et al., 2017) and word tags with natural language texts (Chang et al., 2016).

We only listed a subset of prominent explanation methods, for a more detailed and comprehensive treatment we refer to surveys from Adadi and Berrada (2018), Guidotti et al. (2018), Ras et al. (2018) and Zhang and Chen (2018). In terms of explanation methods, the majority of methods use feature attribution as explanations through saliency methods (eg. Chen et al., 2016; Montavon et al., 2018; Zhang et al., 2018) as well as feature importance values (Lundberg and Lee, 2017). Example-based explanations generate their examples differently, eg. as class prototypes (Li et al., 2017) or most influential training data (Koh and Liang, 2017). Explanations of model internals include, for example, presenting decision rules from decision trees (Lim et al., 2009) or visualizing neurons of a neural network (Olah et al., 2018). Surrogate models from methods such as LIME (Ribeiro et al., 2016) can be explained using other explanation methods, ie. feature attribution, example-based, and model internals. For example, given a decision tree is chosen as surrogate model, it can be explained by feature attribution.

Implicit information collection is common in recommender systems (Chang et al., 2016; Chen et al., 2018; Quijano-Sanchez et al., 2017; Zhang et al., 2014). On the other hand, interactive explanation interfaces (Olah et al., 2018; Sokol and Flach, 2018) incorporate manual adjustments by the explainee, which may be seen as explicit information collection. Other works in ML explanations used feedback from human subjects to improve explanation interpretability. For example, using explanations from experts as an additional model constraint (Ross et al., 2017). Lage et al. (2018) asked a user to evaluate multiple models for explanations. Users were supposed to learn to solve the task, ie. image classification based on the explanations provided. The model giving the explanations that resulted in highest prediction accuracy by users was chosen.

Multiple works mention some form of personalization. Group personalization is mostly done for only one member of a category, for example, evaluating the method on non-experts on ML (Ribeiro et al., 2016) or domain experts (Wu et al., 2018). In contrast, Lim et al. (2009) differentiates explainees by their prior knowledge on the explanation method, while Quijano-Sanchez et al. (2017) use personality as one of their explainee categories. As for those which addressed individual personalization, they are either: i) allowing only manual personalization or ii) they are recommender systems explanations. Manual personalization is enabled either via explanation interfaces (Olah et al., 2018a; Sokol and Flach, 2018) or allowing the incorporation of Bayesian priors (Wang et al., 2016). Explanations for recommender systems are often personalized due to the nature of the task (and the training data). For example, reviews from individuals (Zhang et al., 2014) or browsing activities (Chang et al., 2016) serves as training data for the ML model as well as explainee data used for explanations.

To summarize, a systematic treatment of personalized explanations on a conceptual level is absent. Furthermore, existing methods do not cover the entire design space of existing explanation methods, eg. automatic personalization techniques using explicit information collection have not been developed for ML in general for both individuals and categories. This is despite the acknowledgement that an explanation's interpretability may vary considerably between explainees (Kirsch, 2017; Ras et al., 2018; Tomsett et al., 2018).

## 5    Explainee Data and Information

We describe what data and information can be collected on an explainee and how to obtain it.

### 5.1    Kinds of Explainee Information

Our synthesis of work on personalization and explanation in ML yielded four categories of explainee data: i) *prior knowledge* – what an explainee knows, ii) *decision information* – what information an explainee uses for decision making, iii) *preferences* – what an explainee likes and prefers, and iv) *purpose* – what the explanation is used for.

Prior knowledge is partitioned into ML knowledge and task domain knowledge:

**Machine Learning Knowledge** refers to the explainee's expertise regarding the ML method to be explained. One might distinguish based on explainee roles such as ML engineers or end users, implying a certain level of knowledge (Ras et al., 2018). For example, different levels of deep neural network (DNN) knowledge are described in (Ras et al., 2018): i) Knowledge about detailed mathematical theories and DNNs principles, ii) knowledge to train and integrate DNN models into a final application, and iii) no DNN knowledge, ie. neither theories nor implementation.

**Task Domain Knowledge** refers to the users' domain knowledge on the task at hand, eg. a doctor or a patient using a disease recognition system (Doshi-Velez and Kim, 2017). The nature of the user's expertise influences what level of sophistication they expect in their explanations. For example, domain experts may prefer a somewhat larger and sophisticated model—which confirms facts they know—over a smaller, more opaque one (Lavrač, 1999).

**Decision information** refers to the information utilized by an explainee when she performs the ML task. The information might be rooted in domain knowledge but it might also stem from ML or general knowledge or experiences. It is one of the most important criteria for interpretability of explanations (Miller, 2018), namely coherence with prior beliefs. For example, for an image recognition task decision information could be which parts of the image the user would have used to classify that image.

Decision information might be gathered at different levels, ie. training data and feature level. One might ask an explainee what data samples justify his explanations the most. For example, two doctors might have been exposed to the same patient information during their training, but they might assign different relevance to patient cases. Extracted features from the data by an individual (and their importance) for decision making also constitutes decision information. For instance, a person might deem colors of objects more important than shapes in image recognition.

Decision information might be collected before any explanation have been made or afterwards, eg. using feedback or answers to questions on the computed presentations might be used in an iterative manner to improve explanations (Fails and Olsen Jr, 2003). As the explanation better represents the user's prior beliefs, it may become less faithful to the model, ie. fidelity might decrease. Still, there might be multiple possible explanations which are further filtered by individuals (Miller, 2018) and might lead to similar fidelity.

**Preferences** refer to a subjective prioritization of options of an explainee that are not necessarily relevant for any objective measure of the explanation quality such as objective interpretability. But they might strongly impact a person's feeling towards an explanation or her acceptance (plausibility). Preferences include, for example, information such as user's importance rating or constraints with respect to the desiderata of the model (eg. level of privacy a user wishes to maintain), the desired level of detail of the explanation, the time or effort a user wants to invest to obtain in understanding the explanation, presentation form and employed line of reasoning (eg. explaining based on prototypes or using general rules).

**Purpose** refers to the intended use of an explanation. Prior work on explanation has derived purpose based on functional roles (Ras et al., 2018; Tomsett et al., 2018), eg. end user, developer or data subject. Lipton (2017) lists reasons why ML interpretability is desired, namely *trust*, *causality*, *transferability*, *informativeness*, and *fair and ethical decision making*. The purpose might also be seen as obtaining an answer to explanatory questions, ie. what, how and why (Miller, 2018). Explainable artificial intelligence presented in (Adadi and Berrada, 2018) mentions justify, control, improve and discover. With respect to personalization, we add the goal of *persuasion*, which aims at changing someone's beliefs through reasoning and argument. Persuasion is a common theme in recommender systems (Cremonesi et al., 2012).

**Further information** that can be found in the literature on personalization such as "prior experience with the system" might also be utilized.

### 5.2 Obtaining Explainee Data and Information

Preferences can be elucidated using multiple techniques ranging from traditional means to computer aided methods (Chen and Pu, 2004). Extracting ML and task domain knowledge as well as decision information can be done using knowledge extraction methods (Hoffman et al., 1995; Liou, 1992). The techniques differ strongly based on whether the required information is tacit or explicit. A person might find it difficult to express tacit knowledge, eg. to identify what she utilizes in her decision process. But it might be relatively easy for a person to judge whether she is an expert on a topic or not (explicit knowledge) or state the purpose of the explanation. Elicitation techniques empirically analyzed in (Dieste and Juristo, 2011) are classified based on (Hoffman et al., 1995) into three categories as follows:

- *Analysis of familiar tasks* – investigating what explainees do when performing tasks under their usual condition, eg. protocol analysis, unobtrusive observation, or simulated task
- *Interviews* – asking explainees directly, eg. using unstructured or structured interviews
- *Contrived techniques* – investigating what explainees do when performing modified tasks, eg. scaling, sorting, or hierarchical structuring. An exemplary technique is the repertory grid (McGeorge and Rugg, 1992). It can be applied to ML problems such as classification. The goal is to elucidate information on how people classify elements by first selecting a group of elements representing a relevant aspect of the domain. Significant constructs are identified by presenting three elements and asking how two are similar and thus different from the third.

Interviews are suitable to elicit explicit knowledge. Eliciting tacit knowledge of an explainee can be done with ML techniques (Webb et al., 2001). In this case, analysing familiar tasks and contrived techniques can be used to classify these ML techniques as shown in Table 2.

| Category | Exemplary Works |
|---|---|
| Analysis of familiar tasks | Usage log, eg. from web (Geng and Tian, 2015), browsers (Chang et al., 2016) based on mouse cursor traces (Schneider et al, 2017) or executed commands (Damevski et al, 2017) |
| | Protocol analysis, eg. sharpening important parts of an image (Das et al., 2017) |
| Contrived techniques | Sorting, eg. assigning document relevance (Maddalena et al, 2016), annotating video (Prest et al, 2012) or ML expl. (Ross et al., 2017), selecting images of a concept (Kim et al., 2018) |

*Table 2.      Exemplary ML techniques to elicit tacit knowledge*

Hoffman et al. (1995) put a strong emphasis on knowledge elicitation from experts. Other (overlapping) categorizations have been derived for related tasks. For instance, to learn user profiles (Montaner et al., 2003) distinguish three categories in the context of recommender systems: i) manual – the explainee explicitly states the required information, ii) stereotyping – collecting explainee data based on group memberships, and iii) training set – collecting explainee's logs of tasks relevant to the explained ML model.

Collecting, processing and storing explainee data raises privacy concerns. They must comply to data protection regulations, eg. the European parliament's GDPR. It is crucial to acknowledge and handle the diversity of explainees' perceived personalization and privacy trade-offs (Xu et al., 2011, Awad and Krishnan, 2006; Toch et al., 2012). A survey by Awad and Krishnan (2006), for example, shows a paradox where consumers who are less willing to provide personal information value information transparency the most. Toch et al. (2012) proposed a framework that depicts this trade-off regarding the degree of users' privacy control.

## 6 Personalized Explanation Methods

Personalized explanations can be created using a two-step process, namely: i) customizing explanation properties and ii) generating the personalized explanation.

### 6.1 Customizing explanation properties

A personalized explanation is obtained by adjusting explanation properties making use of explainee data and taking into account the explanation method as illustrated in Table 3. We first discuss how explainee data affects each explanation property. Then, we elaborate on how to adjust to the explanation method.

Explainee data relevant to personalize an explanation's **complexity** are ML knowledge, task domain knowledge, cognitive ability and effort willing to spend on explanation. More complex explanations may suit more knowledgeable explainees willing to spend more time on explanations.

**Prioritization of decision information** can be done, eg. according to the explainee's task domain knowledge. Using a disease recognition task as an example, doctors might understand more medical terms than patients leading to different terms used in the explanation, but even two doctors might use different symptoms to diagnose. A more visual example is given by the saliency map in Figure 1. An attribution technique might highlight certain areas contributing most strongly to the decision, ie. the body of the bird. A personalized method for users focusing on heads might emphasize more the head detailed characteristics of it such as eyes or beak as shown in Figure 1. The visible differences between the two explanations raise the question, whether the input and output behaviour of the ML model and the explainee are aligned (fidelity). If the ML model would not arrive at the same decision using the prioritization of features based on explainee data, ie. primarily relying on the bird's head, the personalized explanation might be inappropriate due to its lack of fidelity.

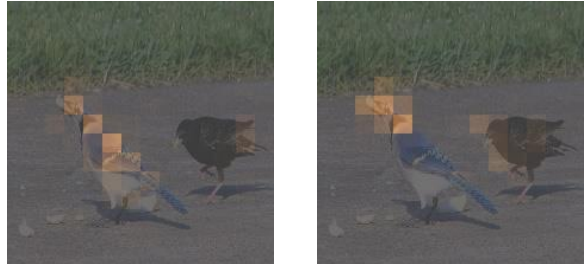

*Figure 1.    Saliency maps for explanation with personalization focusing on heads (right, own figures)*

Explainee data on ML knowledge and purpose are relevant to personalize the **presentation** of an explanation. For example, using an image saliency map highlighting relevant parts for the decision might be enough to convince non-experts of the model's accountability. For ML experts who want to improve the model, however, adding feature visualization might be useful.

|  | Attribution | Example-based | Model Internals | Surrogate Model |
|---|---|---|---|---|
| Comp-lexity | Number of features and/or classes<br><br>Selection of contributor and target, eg. feature-feature or input-output | Number of examples<br><br>Complexity of examples | Size, eg. depth or length<br><br>Feature relationships<br><br>Type of representation, eg. feature visualization | Type of surrogate model, eg. decision tree, decision rules, or linear regression.<br><br>Explanation method of surrogate model |
| Prioritized Decision Information | Features to present | Examples to present<br><br>Features most char-acterizing examples | Features and their relationships to present | Features and their relationships to use and present |
| Presentation | Choice of visualizat-ion technique | Structuring of examples | Choice of visualizat-ion technique | Choice of visualiz-ation technique |

*Table 3.    Exemplary use of explainee data for different explanation methods and properties*

Complexity of **attribution** methods can be personalized by adjusting the number of presented contributors, eg. length of a feature list (sorted by importance), and number of attribution targets, eg. the number of classes to analyze. The relationships between contributors, ie. inputs or lower level features, and attribution target, ie. outputs and higher level features, can also contribute to complexity. For example, it might be difficult to understand relations between features hidden in model internals, as the features extracted by the model might differ to the ones used by the explainee. In contrast, an explainee might be more familiar with inputs and outputs. Moreover, personalization can be done by using relevant features to an explainee in prioritizing decision information, eg. selecting explainee's relevant features in the case of ties on features importance as shown in Figure 1. Presentation is personalized via the selection of visualization technique, eg. highlighting words in sentences or a word list.

An **example-based** explanation may show examples from one or more categories. Categories might be pre-defined such as classes of the input data or based on other information, ie. predicted class probabilities grouped into "low, medium, high". Complexity is personalized by adjusting the number and complexity of examples, eg. their level of detail. To prioritize decision information, ie. to select examples, examples might be taken that best cover features relevant to the explainee. With regards to presentation, examples can be shown with or without structure. An example for a structured presentation is to lay out the examples on a grid based on their similarities as shown in Figure 2, where a feature reduction method, t-SNE (van der Maaten and Hinton, 2008), is used to map the Olivetti face dataset to a two-dimensional grid. Applying such feature reduction methods to features extracted from a ML model, eg. a neural network, allows an explainee to understand better what samples the model considers as similar and dissimilar.

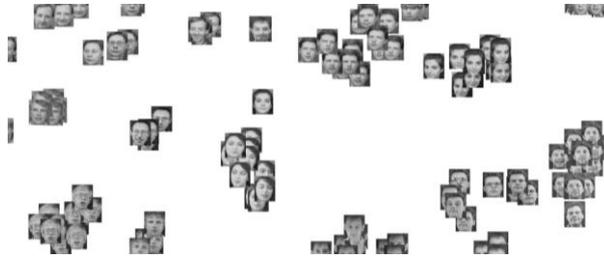

*Figure 2.      t-SNE visualization for faces adapted from van der Maaten and Hinton (2008)*

Personalization of **model internals** depends on its representation. Complexity is personalized by adjusting a model's size, eg. sparseness of a linear model, and features relations, eg. presence of conjunctions in a decision rule, to match the explainee's expertise. The type of representations also contributes to complexity, eg. explaining a neural network by feature visualization or its decision tree approximation. Explainee's relevant features can be prioritized by selecting model internals which involves those features. For example, selecting a decision tree which includes explainee's relevant features or providing visualizations of neurons which corresponds to those features. Another example is to select decision rules which are close to the explainee's decision making rules. Presentation is personalized through selecting appropriate visualization techniques and forms. For example, an explainee may prefer decision rules to be presented in natural language or using imagery as illustrated in Figure 3 (Chen et al., 2018). It is essential to strike a balance between the listed desiderata of personalized explanations such as fidelity and interpretability. Explanations relying primarily on features, which are relevant for an explainee, might lead to high interpretability, but low fidelity.

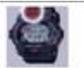

*Figure 3.      Examples of textual and visual explanations adapted from Chen et al. (2018)*

There are multiple **surrogate models** and, thus, the first step in personalization is to choose an appropriate model, eg. for LIME (Ribeiro et al., 2016) the choice is between sparse linear models and decision trees. The selection might depend on the explainee's familiarity with models as well as complexity constraints. Personalization might alter the way a surrogate model is built and, also, how it is explained. For illustration, when a decision tree is chosen as surrogate model, a subset of features, which is used to train the decision tree, can be selected as part of personalization. After the decision tree has been trained, again a subset of features can be selected, which is then presented to the explainee. The choice of the explanation method, ie. attribution, example-based, or model internals, of the surrogate model itself, is also part of personalization.

## 6.2    Post-hoc and intrinsic methods

There are two approaches from interpretable ML, namely, post-hoc and intrinsic explanation, illustrated in Figures 4 and 5, respectively. Both methods might be iterative. That is, an explainee might provide feedback on the provided explanation, which in turn yields an improved explanation.

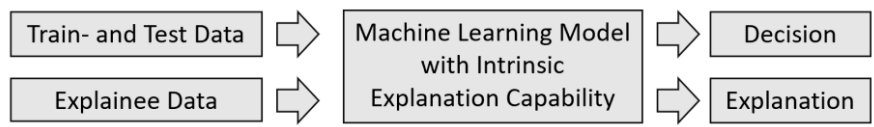

*Figure 4.	In- and outputs of intrinsic personalized explanation methods*

**Intrinsic personalization** seeks to find and train ML models that optimize two goals at the same time, ie. solving the decision task and providing explanation of the decisions (Figure 4). An intrinsic personalized model learns, eg. features relevant to the explainee and of complexity defined by the explainee. Ross et al (2017) use a neural network's explanation as a training constraint. Expert annotated images are used as the 'right reasons' for classifying an image. The differences between the experts' and model's explanations are then minimized. This approach can be personalized by using for each model a separate annotation source. Wu et al. (2018) regularized a neural network by the depth of its decision tree approximations. Apart from constraining features used and degree of complexity, another approach is personalizing the training data. For instance, the ML model might be trained using only data with which an explainee is familiar with. This may cause the ML model to mimic the explainee's decision making process, eg. as in an explainable recommender system (Zhang et al., 2014).

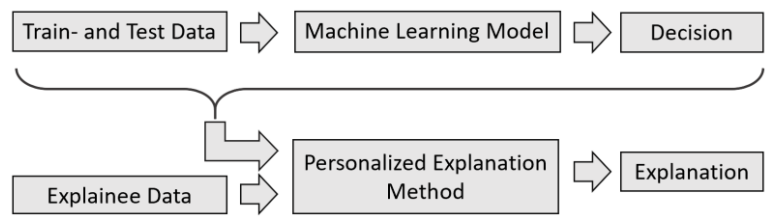

*Figure 5.	In- and outputs of a ML model (top) and a post-hoc personalized explanation method (bottom)*

**Post-hoc personalization** seeks to personalize explanations from trained (non-personalized) ML models. It provides explanations without altering the ML model trained on the original task. It might have access to explainee data that is not used by the ML model solving the task. This approach is applicable to personalize all three explanation aspects, ie. complexity, prioritization of decision information and presentation. For example, in an interface to explain a convolutional neural network (Olah et al., 2018), the explanation methods, eg. feature visualizations or saliency maps, can be adjusted directly by the explainee. Another example relates to decision rules, that is, a decision rule is selected as an explanation among multiple valid decision rules based on its precision (Ribeiro et al., 2018). In terms of personalization, this selection can be based, eg. on the presence of explainee's relevant features in the rule. Another example is presented in (Chen et al., 2018), here an item recommendation is visually explained by marking certain parts on the item's image. They personalized the visual explanations by marking item features which are relevant to the explainee, ie. different explainee with the same recommended item might get different visual explanations.

## 7	Evaluation of personalized explanation methods

Evaluation of desiderata of explanation methods in general is discussed in (Adebayo et al., 2018; Guidotti et al., 2018; Doshi-Velez and Kim, 2017; Ras et al., 2018; Ribeiro et al., 2016, 2018). Though existing work captures useful concepts for evaluation. Overall they are fairly sparse, ie. Adadi and Berrada (2018) found in their review that only 5% of all papers deal with evaluating and quantifying explainability methods. The operationalization of these concepts in terms of mathematical ways of measuring is also underexplored. Here we focus on modifications of existing concepts towards personalization as well as on criteria that only apply for personalization.

**Fidelity** can be evaluated by measuring how well the explanation represent the black-box model's behaviour. For example, when using a transparent model as an explanation, fidelity can be evaluated by measuring the agreement between the transparent model's and black-box model's predictions (Guidotti et al., 2018; Lakkaraju et al., 2017). Evaluating fidelity of a non-personalized and personalized explanations for a particular explainee is identical. Additionally, a comparison between fidelity and interpretability measurements of a non-personalized and personalized explanations helps to evaluate the trade-off in choosing one explanation over the other.

An important aspect of **generalizability** is the range of explainee data a personalization method can incorporate, eg. whether it can include both user's ML knowledge and task domain expertise or not.

As for evaluating **interpretability**, two kinds of measures are formulated in (Fürnkranz et al., 2018): i) an objective measure of the explanation's capability to aid the explainee in performing a task (comprehensibility) and ii) a subjective measure of the explainee's acceptance of the explanation (plausibility). The objective part can be evaluated through measuring explainee performance, eg. the response time and accuracy in simulating a model's decision (Lage et al., 2018; Narayanan et al., 2018; Poursabzi-Sangdeh et al., 2018; Ribeiro et al., 2018). Another objective measure would be measuring the explainee's understanding about the explained model. For example, whether explainees can identify incorrect predictions of a model (Poursabzi-Sangdeh et al., 2018). Whereas the subjective measure is often evaluated through explainee rating on aspects such as plausibility, usefulness, surprisingness, non-triviality, trustworthiness, or overall satisfaction (Fürnkranz et al., 2018; Narayanan et al., 2018; Paulheim, 2012). Explainee's acceptance can also be evaluated indirectly. In Poursabzi-Sangdeh et al. (2018) trust is evaluated by measuring the differences between the model's and explainee's predictions. Another approach is to use the proportion of predictions made by explainees after receiving the explanations, ie. if they were confident enough to make a prediction or not (Ribeiro et al., 2018). A taxonomy of interpretability evaluation proposed in Doshi-Velez and Kim (2017) consists of application-grounded, human-grounded and functionally-grounded evaluation.

**Effort** can be quantified using time the explainee requires to provide the information and to make sense of an explanation. Cognitive load of the explainee to provide information might also be assessed, eg. using eye-tracking systems (Buettner, 2013) or using mental effort ratings and performance scores (Paas and van Merriënboer, 1993).

**Privacy** can be measured using a variety of measures often formulated in terms of what an adversary targeting to obtain confidential information can actually achieve (Wagner and Eckhoff, 2018). It depends on the adversarial model, eg. whether an adversary has access only to information stored permanently (eg. on disc) or to temporary information (eg. in main memory) or whether she can only listen to communication (eg. Internet traffic) or listen to all keystrokes and mouse movements of a user (Goodrich and Tamassia, 2011). Suitable output measures in our context might be information gain, ie. how much an adversary learns about the explainee, or uncertainty, ie. the size of the crowd from which an individual cannot be distinguished from.

**Fairness** can be assessed by computing the distribution of the prior metrics (fidelity, generalizability etc.) across explainees. Larger spread of the distribution is an indicator for low fairness. More complex methods from fairness in machine learning (Kusner et al, 2017) might also be adapted.

## 8     Conclusion

Developing explainee-centric methods for explaining ML models might enhance interpretability. Better explanations are not just relevant due to legislation or to improve existing ML models, but also to create new possibilities for emerging fields such as "machine teaching", ie. machines teaching humans.

Despite efforts towards this goal, our review has indicated significant gaps. We discovered that explicit information collection from explainee is rarely done. Furthermore, our conceptualization revealed that personalized explanation methods differ from conventional explanation methods in aspects such as taking into account an explainee's privacy and effort for collecting data from explainees.